                                                       \newcommand{\Bem}[1]{}

\documentstyle[here,twoside,cite,beispiel]{article}
\title{%
Distribution of degrees of freedom over
structure and motion of rigid bodies %
}
\author{Mieczyslaw A. K{\l}opotek\\
{\footnotesize\sl Institute of Computer Science, 
Polish Academy of Sciences}\\
{\footnotesize\sl 
e-mail:
klopotek{@}ipipan.waw.pl}}

\date{}

\newcommand{\nn}{\nonumber}
\newcommand{\la}{ & &     }
\newcommand{\Zitat}[1]{~\cite{#1}}

\newcommand{\rFig}[1]{fig.~\ref{#1}}
\newcommand{\rTab}[1]{tab.~\ref{#1}}
\newcommand{\rEq}[1]{(\ref{#1})}

\begin{document}

\setcounter{page}{83}
\pagestyle{myheadings}
\maketitle

{
\footnotesize
{\bf Abstract.}\ \
This paper is concerned with recovery of motion and structure parameters
from multiframes under orthogonal projection when only points are traced. 
The main question is how many points and/or how many frames are necessary for
the task. It is demonstrated that 3 frames and 3 points are the absolute
minimum. Closed-form solution is presented. Furthermore, it is shown that the
task may be linearized if either four points or four frames are available. 
It is demonstrated that no increase in the number of points may lead to
recovery of structure and motion parameters from two frames only. It is shown
that instead the increase in the number of points may support the task of
tracing the points from frame to frame.
%
}

\section{Introduction}

Recovery of a three-dimensional structure  from a single view  of even the
simplest scene consisting
of a single object seems to be next to impossible. A number of additional
assumptions seems to be necessary for a successful recovery. Some of clues
mentioned below (and many other) or their combinations proved helpful in the
past:
\begin{itemize}
\item model restrictions (the object belongs to one of parametric classes to
be identified),
\item surface property assumptions (shades, texture etc.), \Zitat{Tsai:1,Tsai:2},
\item usage of synchronized pairs of views (stereoscopic images of various
types),
\item usage of longer sequences of non-synchronized frames,
\item assumptions restricting the pattern of motion (e.g. rotation around a
fixed direction etc.),\Zitat{Lee:1,K:2}, etc.
\end{itemize}

This paper is concerned with recovery of motion and structure parameters
from multiframes under orthogonal projection when only points are traced from
frame to frame (a finite number of them). 
 We assume that the body is rigid that is that the intrinsic (3D) distances
between the traced points are retained from frame to frame. 
We assume further that the only properties we extract from frames
are the distances between projections of traced points (and not for example
their "textures", shades, colors, light reflections etc.). 
We shall call such a body a "point body". 
 We assume that the pattern of motion of the body is
unrestricted between the frames. The term "unrestricted" means that we do not
assume any particular pattern of motion, e.g. rotation around a fixed axis, or
orbiting around an attractive center in gravity field etc., though we
do not forbid such a motion. Still one shall be
concious of the fact that motion of the body reflecting various sides of it is
vital for reconstruction. E.g. the pattern of motion of motionlessness is not
suitable for recovery purposes from multiframes, and also pure shifts of the
body as well as pure rotations around an 
axis orthogonal to the projection plane is not the case, because then
we would have to do with recovery from a single frame.\\
 Though the problem may look fairly simplified, we shall say  that 
similar
problems have already been studied in the past, e.g. \Zitat{Lee:1},  has
been concerned with bodies consisting of two traceable point  rotating
around a fixed 
direction. On the other hand it may be still of practical relevance. Fixing 
traceable points at military vehicles is used to trace the motion of own
troops. In this case the geometry of the rigid body is known and only the
motion may be of interest. But assume the reverse situation. We want to trace
the enemy troops where we only know that vehicles are marked but the geometry
of marking is not known. Here we will have then to do with the complete
problem of recovery of both structure and motion. 
The main question is how many points and/or how many frames are necessary for
the task.
In this paper,
 it is demonstrated that 3 frames and 3 points are the absolute
minimum. 
First, a 
closed-form solution is presented in section 2. In sections 3 and 4 it is
shown that
the task may be linearized if either four points or four frames are available.
In section 5 we are concerned with the problem what kind of information may
be gained if only two
frames are considered. It is demonstrated that no increase in the number of points
may lead to
recovery of structure and motion parameters from two frames only. It is shown,
however, that instead the increase in the number of points may support the
task of tracing the points from frame to frame. 
The paper ends with a brief discussion and some concluding remarks.

\section{Three points and three frames }
    It is an interesting question to investigate the possibility of 
reconstruction of structure and motion from multiframes under orthogonal 
projection. As mentioned in~\Zitat{K:3}, it is possible to recover them from 
three traceable points and three images having a quadratic equation system, 
which may be simplified to a linear one if four images or four frames are
available. 
\subsection{Degrees of freedom for orthogonal projection}

  Each point of the body introduces 3 df in the 
first frame   minus one df for the whole body as there exists 
no possibility of determining the initial depth of the
 body in the space. The 
motion introduces for each subsequent 
frame 5 df only (three for rotations and two for translation), because the
motion in the direction orthogonal to the
projection plane has no impact on the image. In general, with $p$ points forming
the rigid body traced over $k$ frames we have 
$-1+3*p+5*(k-1)$ degrees of freedom.

On the other hand, within each image each traced point provides us with two 
pieces of information: its x and its y position within the frame. Hence 
we have at most 
$ k*2*p$ pieces of information available from k images.
    Thus we need at least to have the balance 
$           -1+3*p+5*(k-1)      \le   k*2*p$
to achieve recoverability. 

    Let us consider some combinations of parameters:
\begin{itemize}
\item for $k=3$ frames, $p=3$ points we get
$-1+3*p+5*(k-1)=18 = k*2*p=18$ 
\item for $k=2$ frames, $p=4$ points we get
$-1+3*p+5*(k-1)=-1+12+5=16 =  k*2*p=2*2*4=16$. 
\end{itemize}
\subsection{Structure and motion for 3 point correspondences}

Let us briefly sketch the procedure of recovery of a three-point
structure from multiframes. 

    Let $P,\, Q,\, R$ be 
the traced points of a rigid body, and $P_i,\, Q_I,\, R_i$ their 
respective projections within the ${i}^{th}$ frame. Let $a,\, b,\, c,\,
a_i,\, b_i,\, c_i$ 
denote the lengths of straight line segments 
$PQ,\, QR,\, RP,\, P_iQ_i,\, Q_iR_i,\, R_iP_i$, respectively. Then for each
frame one of 
the following relationships holds: 
Either: 
$ \sqrt{{a}^2-{a_i}^2}+\sqrt{{b}^2-{b_i}^2}-\sqrt{{c}^2-{c_i}^2}=0$ 
or 
$  \sqrt{{a}^2-{a_i}^2}-\sqrt{{b}^2-{b_i}^2}+\sqrt{{c}^2-{c_i}^2}=0$ 
or
$-\sqrt{{a}^2-{a_i}^2}+\sqrt{{b}^2-{b_i}^2}+\sqrt{{c}^2-{c_i}^2}=0$
(which is easily 
seen from geometrical relationships, 
presented analytically and graphically by K{\l}opotek\Zitat{Klopotek:92g}).
 So
we have three
equations, for $i=1,\, 2,$ and $3$, in three unknowns, $a,\, b,\, c$. As any of the above 
relationships gives after a twofold squaring:

\begin{eqnarray} 
\la
{a}^4+{b}^4+{c}^4-2{a}^2{b}^2-2{a}^2{c}^2-2{b}^2{c}^2 
+
{a_i}^4+{b_i}^4+{c_i}^4-2{a_i}^2{b_i}^2-2{a_i}^2{c_i}^2-2{b_i}^2{c_i}^2
\nn\\
\la
   +2(-{a_i}^2+{b_i}^2+{c_i}^2) {a}^2  
   +2(+{a_i}^2-{b_i}^2+{c_i}^2) {b}^2 \label{p3f3sqr} 
   +2(+{a_i}^2+{b_i}^2-{c_i}^2) {c}^2 = 0
\end{eqnarray}
\\
which is quadratic in ${a}^2,\, {b}^2,\, {c}^2$, hence solvable by exploitation of 
proper methods. 

In an experiment we used a partial linearization approach. 
From formulas for $i=1$ and $i=2$ subtracted with one for $i=3$:


\newcommand{\faca}[1]{ 2(-{a_#1}^2+{b_#1}^2+{c_#1}^2) }
\newcommand{\facb}[1]{ 2( {a_#1}^2-{b_#1}^2+{c_#1}^2) }
\newcommand{\facc}[1]{ 2( {a_#1}^2+{b_#1}^2-{c_#1}^2) }
\newcommand{\facCst}[1]{
({a_#1}^4+{b_#1}^4+{c_#1}^4-2{a_#1}^2{b_#1}^2
-2{a_#1}^2{c_#1}^2-2{b_#1}^2{c_#1}^2) }

%
$$(\faca{1}-\faca{3}){a}^2 +(\facb{1}$$
$$-\facb{3}){b}^2 + (\facc{1}-\facc{3}){c}^2 $$
$$(\facCst{1}$$
$$-\facCst{3}) =0,$$
$$(\faca{2}-\faca{3}){a}^2 +
(\facb{2} $$
$$- \facb{3}){b}^2 
(\facc{2}-\facc{3}){c}^2 $$
$$
(\facCst{2}$$
$$
-\facCst{3}) =0,$$

\newcommand{\diffa}[1]{d_{a,#1}}
\newcommand{\diffb}[1]{d_{b,#1}}
\newcommand{\diffc}[1]{d_{c,#1}}
\newcommand{\diffCst}[1]{d_{Cst,#1}}

\noindent
denoting 
$$\diffa{1} = (\faca{1}-\faca{3}), $$
$$
\diffb{1} =(\facb{1}-\facb{3}),$$
$$
\diffc{1} =(\facc{1}-\facc{3}),$$
$$
\diffCst{1} = (\facCst{1},$$
$$
-\facCst{3}) ,$$
$$
\diffa{2} =(\faca{2}-\faca{3}),$$
$$
\diffb{2} =(\facb{2}-\facb{3}),$$
$$
\diffc{2} =(\facc{2}-\facc{3}),$$
$$
\diffCst{2} =(\facCst{2}\nn,$$
$$
-\facCst{3}),$$

\noindent
 we calculated the
quantity ${a}^2$ and ${b}^2$ as follows:
$$
{a}^2  =  
\frac{(-\diffc{1}{c}^2-\diffCst{1})  \cdot\diffb{2}
     -(-\diffc{2}{c}^2-\diffCst{2})          \cdot\diffb{1}
     }
     { \diffa{1}\cdot\diffb{2}- \diffa{2}\cdot\diffb{1}
     },
$$
$${b}^2  =
\frac{ \diffa{1}\cdot(-\diffc{2}{c}^2-\diffCst{2})
                - \diffa{2}\cdot(-\diffc{1}{c}^2-\diffCst{1})
     }
     { \diffa{1}\cdot\diffb{2}- \diffa{2}\cdot\diffb{1}
     }.
$$

\noindent
Let us introduce notation:

\newcommand{\aonc}{
\frac{(-\diffc{1})  \cdot\diffb{2}
     -(-\diffc{2})          \cdot\diffb{1}
     }
     { \diffa{1}\cdot\diffb{2}- \diffa{2}\cdot\diffb{1}
     }
}
\newcommand{\aonCst}{
\frac{(-\diffCst{1})  \cdot\diffb{2}
     -(-\diffCst{2})  \cdot\diffb{1}
     }
     { \diffa{1}\cdot\diffb{2}- \diffa{2}\cdot\diffb{1}
     }
}
\newcommand{\bonc}{
\frac{ \diffa{1}\cdot(-\diffc{2})
                - \diffa{2}\cdot(-\diffc{1})
     }
     { \diffa{1}\cdot\diffb{2}- \diffa{2}\cdot\diffb{1}
     }
}
\newcommand{\bonCst}{
\frac{ \diffa{1}\cdot(-\diffCst{2})
                - \diffa{2}\cdot(-\diffCst{1})
     }
     { \diffa{1}\cdot\diffb{2}- \diffa{2}\cdot\diffb{1}
     }
}

\newcommand{\Aaonc}{A_c}
\newcommand{\AaonCst}{A_{Cst}}
\newcommand{\Abonc}{B_c}
\newcommand{\AbonCst}{B_{Cst}}
$$ \Aaonc  =\aonc , \quad 
\AaonCst=\aonCst , $$
$$
\Abonc  =\bonc ,\quad  
\AbonCst=\bonCst . $$

\noindent
So we have simply:

$${a}^2 = \Aaonc {c}^2  + \AaonCst ,\quad 
{b}^2 = \Abonc {c}^2  + \AbonCst .
$$

We can now substitute these expressions into the equation for $i=3$:

$$
(A_cc^2+A_{Cst})^2 + (B_cc^2+ B_{Cst})^2 + c^4 - 2(A_cc^2 + A_{Cst})(B_cc^2
+B_{Cst}) $$
$$ -2(A_cc^2 +A_{Cst})c^2 
 - 2(B_cc^2+B_{Cst})c^2
+ (a_3^4 + b_3^4 + c_3^4 - 2a_3^2b_3^2 - 2 a_3^2c_3^2 - 2b_3^2c_3^2)
$$
$$
+2(-a_3^2 + b_3^2 + c_3^2)
(A_cc^2 + A_{Cst})
+2(a_3^2 - b_3^2 + c_3^2)(B_cc^2 + B_{Cst})
+2(a_3^2 + b_3^2 - c _3^2)c^2 = 0. 
$$

It is immediately obvious that the above equation is quadratic in ${c}^2$ and
hence solvable by elementary methods. Then $a$, $b$ and $c$ can be calculated
from previous equations and by square-rooting.

A few comments are necessary at this point. The above equation (and hence
the original problem, as other  variables are uniquely determined by $c$) may
have none,
 one or two solutions (or infinitely many - if the three points happen to be
collinear or two frames prove to be identical up to rotation). No solution may
be attributed to some measurement
errors (or to the fact that the three traced points do not in fact constitute 
a rigid body). The definite solutions need to be checked on physical
feasibility, that is:  
\begin{itemize}
\item
neither ${c}^2$ nor  ${a}^2$ nor  ${b}^2$ can be negative,
\item
neither ${c}^2$ nor  ${a}^2$ nor  ${b}^2$ can be shorter than their respective
projections in frames 1,2 and 3.
\end{itemize}

\newcommand{\DREIECK}[3]
{%
\setlength\unitlength{1cm}
 \begin{picture}(4,4)
 \put(#1){\special{em:moveto} }  
 \put(#2){\special{em:lineto} }  
 \put(#3){\special{em:lineto} }  
 \put(#1){\special{em:lineto} }  

 \put(#1){P}
 \put(#2){Q}
 \put(#3){R}
 \end{picture}
}%


\begin{Bsp}
The  3 point rigid body with geometry    given 
in \rTab{bspp3f3tab}
 has been rotated in space.  
\begin{table}[H]
\label{bspp3f3tab}
\begin{center}
\begin{tabular}{rrrr}
\hline
edge: 
             &                 PQ (a)   &    QR (b)  &      RP (c) \\
  length:        real &         2    &     3  &   4     \\
  length:     squared &         4    &     9  &   16    \\
\hline
\end{tabular}
\end{center}
\caption{Distances   of points of a 3 point rigid body.}
\end{table}

Three projections of that body are shown in \rFig{bspp3f3fig}.        

Two solutions proved to be feasible:
\begin{itemize}
\item ${a}^2$= 4,       ${b}^2$= 9,       ${c}^2$= 16  and,
\item ${a}^2$= 2.6849, ${b}^2$= 8.33902, ${c}^2$= 16.2189.
\end{itemize}

We made also a study of impact of  measurement errors. \\
Assuming errors of up to 0.1 \% we got e.g. 
two solution:\\
a$^2=$ 4.00152, b$^2=$ 8.98252, c$^2=$ 16.009, and 
a$^2=$ 0.8868, b$^2=$ 3.69478, c$^2=$ 13.9678. \\
The first solution approximates the correct solution well (error below 0.2\%).\\
Assuming errors of up to 1. \% we got e.g. 
two solution:\\
a= 4.25684, b= 9.10636, c= 15.9228, and 
a= 1.09062, b= 6.57993, c= 12.2893. \\
The first solution approximates the correct solution not too well (error
below 5\%).\\
\begin{figure}[H]
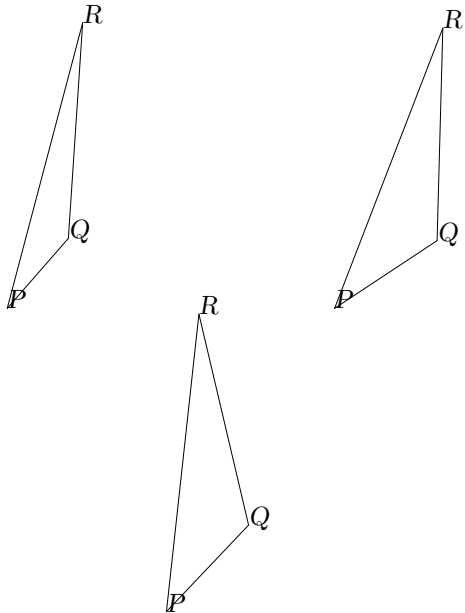

\label{bspp3f3fig}
\centering
%
%
%
%
\DREIECK{ 0.00, 0.00}{ 0.81, 0.93}{ 1.00, 3.80}
%
%
\DREIECK{ 0.00, 0.00}{ 1.36, 0.90}{ 1.44, 3.73}
%
%
\DREIECK{ 0.00, 0.00}{ 1.09, 1.15}{ 0.43, 3.96}
\caption{Three projections of a three point rigid body.}
\end{figure}
Assuming errors of up to 10. \% we got e.g. 
two solution:\\
a= 5.11852, b= 10.8971, c= 15.3443, and 
a= 3.18443, b= 5.14897, c= 13.0865.  
The first solution deviates more than 15 \% from the correct solution.
\end{Bsp}


\section{Three points and four frames }

    K{\l}opotek,\Zitat{Klopotek:92g}, simplified the equation system
\rEq{p3f3sqr} for 3 
traceable points by 
using four instead of three frames and subtracting the twofold squared 
equation for the first 
frame from those of the other ones. So one obtains three equations of the 
form for i=2, 3, 4: 
$$
 {a_i}^4+{b_i}^4+{c_i}^4-2{a_i}^2{b_i}^2-2{a_i}^2{c_i}^2-2{b_i}^2{c_i}^2-
{a_1}^4-{b_1}^4-{c_1}^4+2{a_1}^2{b_1}^2+2{a_1}^2{c_1}^2
$$
$$ 
+2{b_1}^2{c_1}^2=2{a}^2( {a_i}^2-{b_i}^2-{c_i}^2-{a_1}^2+{b_1}^2+{c_1}^2)+
 2{b}^2(-{a_i}^2+{b_i}^2-{c_i}^2+{a_1}^2-{b_1}^2  
$$
$$
+{c_1}^2)+  2{c}^2(-{a_i}^2-{b_i}^2+{c_i}^2+{a_1}^2+{b_1}^2-{c_1}^2)
$$
which are linear in  ${a}^2,\, {b}^2,\, {c}^2$, hence solvable by exploitation of 
respective methods. (No linear dependence is introduced as a new frame is 
exploited unless the motion has a very special form.)

It should be noted that compared to the case of three frames with three points
we gain the uniqueness of the solution (previously two solutions could prove
correct for the given frames).

A numerical example can be found in\Zitat{Klopotek:92g}.

In an analysis of measurement errors we assumed the following values of 
a =   2, b = 3, c = 3.562, hence 
a$^2=$ 4, b$^2=$  9, c$^2=$  12.6878.\\
Assuming error level of up to 0.1\% we got e.g.: 
 a$^2=$  4.00805,
 b$^2=$  8.99673,
 c$^2=$  12.6841\\
which seems to be quite satisfactory.\\
Assuming error level of up to 1.\% we got e.g.:
 a$^2=$  4.08991,
 b$^2=$  8.99485,
 c$^2=$  12.6539\\
which is not bad.\\
Assuming error level of up to 10.\% we got the worst case:
 a$^2=$   1.02533,
 b$^2=$   3.78813,
 c$^2=$   9.84095 
which is desastrous. However, the average performance was with 20\% 
from deviation of
correct values.

\section{Four points and three frames }

    Let $P$, $Q$, $R$, $T$ be the traced points of a rigid body, and 
$P_i$, $Q_I$, $R_i$, $T_i$ their 
respective projections within the ${i}^{th}$ frame. Let $a$, $b$, $c$, $d$, 
$g$, $f$, $a_i$, $b_i$, $c_i$, $d_i$, $g_i$, $f_i$ 
denote the lengths of straight line segments 
$PQ$, $QR$, $RP$, $TR$, $TQ$, $TP$, $P_iQ_i$,  
$Q_iR_i$, $R_iP_i$, $T_iR_i$, $T_iQ_i$, $T_iP_i$, 
respectively. Then for each frame three relationships hold:
\begin{eqnarray}
\la
 \sqrt{{g}^2-{g_i}^2}=\pm\sqrt{{a}^2-{a_i}^2}\pm\sqrt{{f}^2-{f_i}^2},
\nn
\end{eqnarray}
and
\begin{eqnarray} \label{p4f2}
\la
 \sqrt{{d}^2-{d_i}^2}=\pm\sqrt{{b}^2-{b_i}^2}\pm\sqrt{{g}^2-{g_i}^2},
\end{eqnarray}
and
\begin{eqnarray}
\la
 \sqrt{{f}^2-{f_i}^2}=\pm\sqrt{{c}^2-{c_i}^2}\pm\sqrt{{d}^2-{d_i}^2}.
\nn
\end{eqnarray}
Then from \rEq{p4f2} we obtain the linear equation 
system for i=2, 3:\\
$${a_i}^4+{g_i}^4+{f_i}^4-2{a_i}^2{g_i}^2-2{a_i}^2{f_i}^2-2{g_i}^2{f_i}^2-
{a_1}^4-{g_1}^4-{f_1}^4+2{a_1}^2{g_1}^2+2{a_1}^2{f_1}^2
$$
$$
+2{g_1}^2{f_1}^2 =2{a}^2( {a_i}^2-{g_i}^2-{f_i}^2-{a_1}^2
+{g_1}^2+{f_1}^2)+ 2{g}^2(-{a_i}^2+{g_i}^2-{f_i}^2+{a_1}^2
$$
$$
-{g_1}^2+{f_1}^2)+  2{f}^2(-{a_i}^2-{g_i}^2+{f_i}^2+{a_1}^2+{g_1}^2-{f_1}^2), 
$$
and
$$
 {d_i}^4+{b_i}^4+{g_i}^4-2{d_i}^2{b_i}^2-2{d_i}^2{g_i}^2-2{b_i}^2{g_i}^2-
{d_1}^4-{b_1}^4-{g_1}^4+2{d_1}^2{b_1}^2+2{d_1}^2{g_1}^2
$$
$$
+2{b_1}^2{g_1}^2 = 2{d}^2( {d_i}^2-{b_i}^2-{g_i}^2-{d_1}^2+{b_1}^2+{g_1}^2)+
   2{b}^2(-{d_i}^2+{b_i}^2-{g_i}^2
$$
$$
+{d_1}^2-{b_1}^2+{g_1}^2)+
2{g}^2(-{d_i}^2-{b_i}^2+{g_i}^2+{d_1}^2+{b_1}^2-{g_1}^2), 
$$
and
$$
{d_i}^4+{f_i}^4+{c_i}^4-2{d_i}^2{f_i}^2-2{d_i}^2{c_i}^2-2{f_i}^2{c_i}^2- 
{d_1}^4-{f_1}^4-{c_1}^4+2{d_1}^2{f_1}^2+2{d_1}^2{c_1}^2
$$
$$
+2{f_1}^2{c_1}^2= 2{d}^2( {d_i}^2-{f_i}^2-{c_i}^2-{d_1}^2+{f_1}^2+{c_1}^2)+
$$
$$
   2{f}^2(-{d_i}^2+{f_i}^2-{c_i}^2+{d_1}^2-{f_1}^2+{c_1}^2)+ 
 2{c}^2(-{d_i}^2-{f_i}^2+{c_i}^2+{d_1}^2+{f_1}^2-{c_1}^2) 
$$
\\
This linear equation system is easily solved.

Again we obtain always (at most) a single solution instead of two as may be
the case with three frames and three points only.

\newcommand{\DREID}[4]
{%
\setlength\unitlength{1cm}
 \begin{picture}(4,4)
 \put(#1){\special{em:moveto} }  
 \put(#2){\special{em:lineto} }  
 \put(#3){\special{em:lineto} }  
 \put(#1){\special{em:lineto} }  
 \put(#4){\special{em:moveto} }  
 \put(#2){\special{em:lineto} }  
 \put(#4){\special{em:moveto} }  
 \put(#3){\special{em:lineto} }  
 \put(#4){\special{em:moveto} }  
 \put(#1){\special{em:lineto} }  

 \put(#1){P}
 \put(#2){Q}
 \put(#3){R}
 \put(#4){T}

 \end{picture}
}%


\begin{Bsp}
The 4 point rigid body,  geometry of which is given in  \rTab{bspp4f3tab},
 has been rotated in space. 

\begin{table}[H]
\begin{center}
\begin{tabular}{rrrr}
\hline
edge: 
             &                 PQ    &    QR  &      RP \\
  length:        real &         2    &     3  &   3.562 \\
  length:     squared &         4    &     9  & 12.6878 \\
\hline
edge:                 &         TP   &     TQ &       TR\\
  length:         real&    7.07107   &7.43303 &  5.82734 \\
  length:      squared&         50   &  55.25 &  33.9578 \\
\hline
\end{tabular}
\end{center}
\caption{Distances of points in a  4 point rigid body.}
\label{bspp4f3tab}
\end{table}

Three projections are shown in the \rFig{bspp4f3fig}.


We denote edges as:
\\
\begin{center}
\begin{tabular}{|r|r|r|r|r|r|}
\hline 
           PQ & QR & RP & TP & TQ & TR \\
\hline 
          $\sqrt{x1}$  &   $\sqrt{x2}$ &  $\sqrt{x3}$  
&   $\sqrt{x4}$ &          $\sqrt{x5}$ &   $\sqrt{x6}$ \\
\hline 
\end{tabular}
\end{center}
\ \\

Then we obtain the equation system (as a matrix) in variables x1-x6 in 
\rTab{bspp4f3tab2}
\begin{figure}[H]
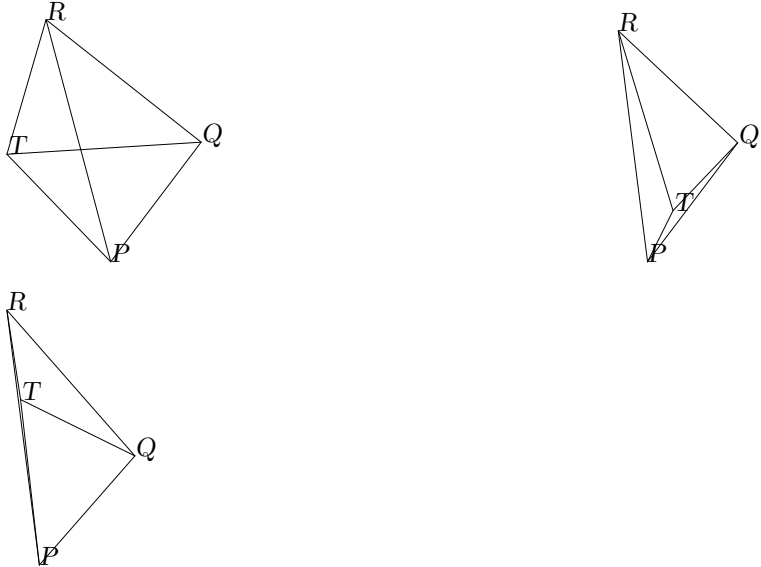

\label{bspp4f3fig}
%
%
%
%
%
\DREID{ 1.38, 0.00}{ 2.58, 1.59}{ 0.52, 3.22}{ 0.00, 1.43}
%
%
%
\DREID{ 0.39, 0.00}{ 1.59, 1.58}{ 0.00, 3.07}{ 0.73, 0.68}
%
%
%
\DREID{ 0.43, 0.00}{ 1.70, 1.45}{ 0.00, 3.39}{ 0.18, 2.20}
\caption{3 projections of a 4 point rigid body.}
\end{figure}

\begin{table}[H] 
\begin{center}
 \begin{tabular}{rrrrr}
 x1   &  x2   &  x3   &  x4   &  x5  \\
\hline  -16.9258  &  0  &  0  &  -3.65832  &  3.43508 \\
 0  &  -0.423132  &  0  &  0  &  11.5049 \\
 0  &  0  &  1.82806  &  9.25371  &  0 \\
 -5.18936  &  0  &  0  &  -9.99644  &  9.04194 \\
 0  &  -11.1085  &  0  &  0  &  3.03494 \\
 0  &  0  &  -3.25323  &  -4.82034  &  0 \\
\end{tabular}
 \begin{tabular}{rr}
 x6   &  1  \\
\hline  0  &  60.831  \\
 -20.161  &  52.7865  \\
 -15.3188  &  34.3132 \\
 0  &  21.0125  \\
 -4.07729  &  70.7523  \\
 7.10581  &  40.9956  \\
\end{tabular}
\end{center}
\caption{The coefficient matrix.} 
\label{bspp4f3tab2}
\end{table}

\noindent
The solution of the above equation system is:\\
 x1 = 4, 
 x2 = 9, 
 x3 = 12.6878,
 x4 = 50,
 x5 = 55.25,
 x6 = 33.9578,\\
which means perfect agreement with the intrinsic rigid body.

The linearity has clearly its price: that is the sensitivity to measurement
errors. If we have random errors of up to 0.1 \% of the real value, then
we get still reasonable results, e.g. in a test run we had:\\
 x1 = 4.00643,
 x2 = 9.16415,
 x3 = 12.8076,
 x4 = 50.6951,
 x5 = 56.0627,
 x6 = 34.4038\\
On average, errors for edge lengths (square roots of the above) did not exceed
1 \%. However  random measurement errors of up to 1. \% of the real value,
lead to serious deterioration of results,
e.g. in a test run we had:\\
 x1 = 4.09934,
 x2 = 11.2053,
 x3 = 14.2781,
 x4 = 58.6152,
 x5 = 65.3898,
 x6 = 39.4439,\\
which means errors of well above 20 \%.
\end{Bsp}

%

\section{Handling 2 frames}

    So let us now consider a rigid body with four points over two frames (i=1,
2).  Let us consider the equation system \rEq{p4f2}. 
   Please notice that we have also a fourth relationship related to the 
triangle ABC:
$\sqrt{{a}^2-{a_i}^2}=\sqrt{{b}^2-{b_i}^2}+\sqrt{{c}^2-{c_i}^2}$
but we make no use of it as it is linearly dependent on the three previous 
ones.

    In this way we got 6 equations (3 for each of the two frames) in six 
variables a, b, c, d, g, f. The respective twofold squaring leads to quadratic 
equations. However, we cannot solve this equation system because, as
we demonstrate below,  
 they are dependent - see next subsection. Thereafter, in a subsection to
follow, we   show how unsolvability of this equation system may be exploited
for point identification problem. The last subsection discusses consequences
for recovery of  curves from two frames.

\subsection{Two frames - insufficient for recovery}

\setlength\unitlength{3cm}      

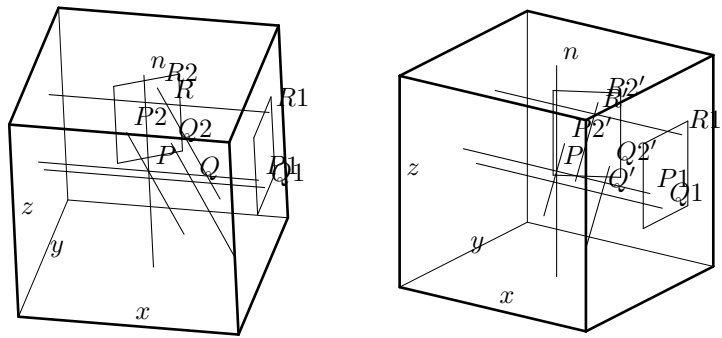
\begin{figure}[H]
\special{em:linewidth 0.4pt}
\begin{picture}(1.73,1.73)

\special{em:linewidth 0.2pt}
\put(0.507,0.737){\special{em:moveto} }
\put(1.482,0.658){\special{em:lineto} }
\put(0.507,0.737){\special{em:moveto} }
\put(0.289,0.220){\special{em:lineto} }
\put(0.507,0.737){\special{em:moveto} }
\put(0.466,1.589){\special{em:lineto} }
\special{em:linewidth 1.0pt}
\put(1.264,0.141){\special{em:moveto} }
\put(0.289,0.220){\special{em:lineto} }
\put(1.264,0.141){\special{em:moveto} }
\put(1.482,0.658){\special{em:lineto} }
\put(1.264,0.141){\special{em:moveto} }
\put(1.223,0.993){\special{em:lineto} }
\put(1.441,1.510){\special{em:moveto} }
\put(0.466,1.589){\special{em:lineto} }
\put(1.441,1.510){\special{em:moveto} }
\put(1.223,0.993){\special{em:lineto} }
\put(1.441,1.510){\special{em:moveto} }
\put(1.482,0.658){\special{em:lineto} }
\put(0.248,1.072){\special{em:moveto} }
\put(1.223,0.993){\special{em:lineto} }
\put(0.248,1.072){\special{em:moveto} }
\put(0.466,1.589){\special{em:lineto} }
\put(0.248,1.072){\special{em:moveto} }
\put(0.289,0.220){\special{em:lineto} }
\put(0.807,0.210){$x$}
\put(0.428,0.509){$y$}
\put(0.298,0.676){$z$}
\special{em:linewidth 0.4pt}

\put(0.895,0.895){$P$}

\put(1.090,0.848){$Q$}

\put(0.975,1.192){$R$}

\put(1.383,0.855){$P1$}

\put(1.409,0.822){$Q1$}

\put(1.429,1.155){$R1$}

\put(0.798,1.067){$P2$}

\put(0.995,1.018){$Q2$}

\put(0.934,1.264){$R2$}

\put(0.874,1.321){$n$}

\put(1.408,1.194){\special{em:moveto} }
\put(1.424,0.853){\special{em:lineto} }

\put(1.424,0.853){\special{em:moveto} }
\put(1.348,0.672){\special{em:lineto} }

\put(1.348,0.672){\special{em:moveto} }
\put(1.331,1.013){\special{em:lineto} }

\put(1.331,1.013){\special{em:moveto} }
\put(1.408,1.194){\special{em:lineto} }

\put(0.998,1.295){\special{em:moveto} }
\put(1.015,0.954){\special{em:lineto} }

\put(1.015,0.954){\special{em:moveto} }
\put(0.727,0.900){\special{em:lineto} }

\put(0.727,0.900){\special{em:moveto} }
\put(0.710,1.241){\special{em:lineto} }

\put(0.710,1.241){\special{em:moveto} }
\put(0.998,1.295){\special{em:lineto} }

\put(0.844,1.291){\special{em:moveto} }
\put(0.886,0.439){\special{em:lineto} }

\put(1.353,0.825){\special{em:moveto} }
\put(0.377,0.905){\special{em:lineto} }

\put(1.379,0.792){\special{em:moveto} }
\put(0.403,0.871){\special{em:lineto} }

\put(1.399,1.125){\special{em:moveto} }
\put(0.424,1.204){\special{em:lineto} }

\put(0.768,1.037){\special{em:moveto} }
\put(1.022,0.585){\special{em:lineto} }

\put(0.965,0.988){\special{em:moveto} }
\put(1.248,0.482){\special{em:lineto} }

\put(0.904,1.234){\special{em:moveto} }
\put(1.181,0.742){\special{em:lineto} }
\end{picture}
%
\special{em:linewidth 0.4pt}
\begin{picture}(1.73,1.73)

\special{em:linewidth 0.2pt}
\put(0.735,0.639){\special{em:moveto} }
\put(1.560,0.442){\special{em:lineto} }
\put(0.735,0.639){\special{em:moveto} }
\put(0.170,0.351){\special{em:lineto} }
\put(0.735,0.639){\special{em:moveto} }
\put(0.735,1.576){\special{em:lineto} }
\special{em:linewidth 1.0pt}
\put(0.995,0.154){\special{em:moveto} }
\put(0.170,0.351){\special{em:lineto} }
\put(0.995,0.154){\special{em:moveto} }
\put(1.560,0.442){\special{em:lineto} }
\put(0.995,0.154){\special{em:moveto} }
\put(0.995,1.091){\special{em:lineto} }
\put(1.560,1.379){\special{em:moveto} }
\put(0.735,1.576){\special{em:lineto} }
\put(1.560,1.379){\special{em:moveto} }
\put(0.995,1.091){\special{em:lineto} }
\put(1.560,1.379){\special{em:moveto} }
\put(1.560,0.442){\special{em:lineto} }
\put(0.170,1.288){\special{em:moveto} }
\put(0.995,1.091){\special{em:lineto} }
\put(0.170,1.288){\special{em:moveto} }
\put(0.735,1.576){\special{em:lineto} }
\put(0.170,1.288){\special{em:moveto} }
\put(0.170,0.351){\special{em:lineto} }
\put(0.613,0.282){$x$}
\put(0.482,0.525){$y$}
\put(0.200,0.849){$z$}
\special{em:linewidth 0.4pt}

\put(0.895,0.895){$P$}

\put(1.094,0.796){$Q'$}

\put(1.065,1.148){$R'$}

\put(1.308,0.796){$P1$}

\put(1.364,0.731){$Q1$}

\put(1.449,1.056){$R1$}

\put(0.930,1.017){$P2'$}

\put(1.129,0.916){$Q2'$}

\put(1.079,1.199){$R2'$}

\put(0.895,1.363){$n$}

\put(1.447,1.087){\special{em:moveto} }
\put(1.447,0.712){\special{em:lineto} }

\put(1.447,0.712){\special{em:moveto} }
\put(1.249,0.611){\special{em:lineto} }

\put(1.249,0.611){\special{em:moveto} }
\put(1.249,0.986){\special{em:lineto} }

\put(1.249,0.986){\special{em:moveto} }
\put(1.447,1.087){\special{em:lineto} }

\put(1.149,1.212){\special{em:moveto} }
\put(1.149,0.837){\special{em:lineto} }

\put(1.149,0.837){\special{em:moveto} }
\put(0.850,0.848){\special{em:lineto} }

\put(0.850,0.848){\special{em:moveto} }
\put(0.850,1.223){\special{em:lineto} }

\put(0.850,1.223){\special{em:moveto} }
\put(1.149,1.212){\special{em:lineto} }

\put(0.865,1.333){\special{em:moveto} }
\put(0.865,0.397){\special{em:lineto} }

\put(1.278,0.766){\special{em:moveto} }
\put(0.452,0.964){\special{em:lineto} }

\put(1.334,0.701){\special{em:moveto} }
\put(0.509,0.899){\special{em:lineto} }

\put(1.419,1.026){\special{em:moveto} }
\put(0.593,1.223){\special{em:lineto} }

\put(0.900,0.987){\special{em:moveto} }
\put(0.808,0.667){\special{em:lineto} }

\put(1.099,0.886){\special{em:moveto} }
\put(0.996,0.529){\special{em:lineto} }

\put(1.049,1.169){\special{em:moveto} }
\put(0.949,0.820){\special{em:lineto} }
\end{picture}
\caption{First match of two views.} \label{zweiframeseins}
\end{figure}

%

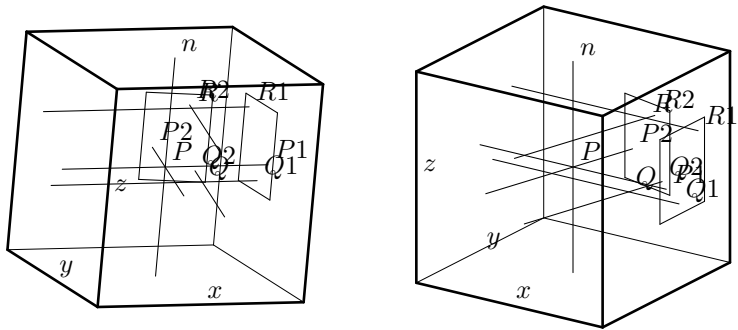
\begin{figure}[H]
\special{em:linewidth 0.4pt}
\begin{picture}(1.73,1.73)

\special{em:linewidth 0.2pt}
\put(1.079,0.519){\special{em:moveto} }
\put(0.166,0.498){\special{em:lineto} }
\put(1.079,0.519){\special{em:moveto} }
\put(1.479,0.265){\special{em:lineto} }
\put(1.079,0.519){\special{em:moveto} }
\put(1.164,1.486){\special{em:lineto} }
\special{em:linewidth 1.0pt}
\put(0.566,0.244){\special{em:moveto} }
\put(1.479,0.265){\special{em:lineto} }
\put(0.566,0.244){\special{em:moveto} }
\put(0.166,0.498){\special{em:lineto} }
\put(0.566,0.244){\special{em:moveto} }
\put(0.651,1.211){\special{em:lineto} }
\put(0.251,1.465){\special{em:moveto} }
\put(1.164,1.486){\special{em:lineto} }
\put(0.251,1.465){\special{em:moveto} }
\put(0.651,1.211){\special{em:lineto} }
\put(0.251,1.465){\special{em:moveto} }
\put(0.166,0.498){\special{em:lineto} }
\put(1.564,1.232){\special{em:moveto} }
\put(0.651,1.211){\special{em:lineto} }
\put(1.564,1.232){\special{em:moveto} }
\put(1.164,1.486){\special{em:lineto} }
\put(1.564,1.232){\special{em:moveto} }
\put(1.479,0.265){\special{em:lineto} }
\put(1.052,0.285){$x$}
\put(0.396,0.401){$y$}
\put(0.639,0.758){$z$}
\special{em:linewidth 0.4pt}

\put(0.895,0.895){$P$}

\put(1.058,0.829){$Q$}

\put(1.012,1.156){$R$}

\put(1.351,0.905){$P1$}

\put(1.303,0.834){$Q1$}

\put(1.268,1.162){$R1$}

\put(0.838,0.983){$P2$}

\put(1.027,0.878){$Q2$}

\put(1.003,1.170){$R2$}

\put(0.938,1.379){$n$}

\put(1.223,1.193){\special{em:moveto} }
\put(1.189,0.806){\special{em:lineto} }

\put(1.189,0.806){\special{em:moveto} }
\put(1.329,0.718){\special{em:lineto} }

\put(1.329,0.718){\special{em:moveto} }
\put(1.363,1.105){\special{em:lineto} }

\put(1.363,1.105){\special{em:moveto} }
\put(1.223,1.193){\special{em:lineto} }

\put(1.076,1.184){\special{em:moveto} }
\put(1.042,0.797){\special{em:lineto} }

\put(1.042,0.797){\special{em:moveto} }
\put(0.747,0.810){\special{em:lineto} }

\put(0.747,0.810){\special{em:moveto} }
\put(0.781,1.197){\special{em:lineto} }

\put(0.781,1.197){\special{em:moveto} }
\put(1.076,1.184){\special{em:lineto} }

\put(0.908,1.349){\special{em:moveto} }
\put(0.822,0.381){\special{em:lineto} }

\put(1.321,0.875){\special{em:moveto} }
\put(0.409,0.855){\special{em:lineto} }

\put(1.273,0.804){\special{em:moveto} }
\put(0.360,0.783){\special{em:lineto} }

\put(1.238,1.132){\special{em:moveto} }
\put(0.326,1.111){\special{em:lineto} }

\put(0.808,0.953){\special{em:moveto} }
\put(0.948,0.736){\special{em:lineto} }

\put(0.997,0.848){\special{em:moveto} }
\put(1.128,0.643){\special{em:lineto} }

\put(0.973,1.140){\special{em:moveto} }
\put(1.107,0.932){\special{em:lineto} }
\end{picture}
%
\special{em:linewidth 0.4pt}
\begin{picture}(1.73,1.73)

\special{em:linewidth 0.2pt}
\put(0.735,0.639){\special{em:moveto} }
\put(1.560,0.442){\special{em:lineto} }
\put(0.735,0.639){\special{em:moveto} }
\put(0.170,0.351){\special{em:lineto} }
\put(0.735,0.639){\special{em:moveto} }
\put(0.735,1.576){\special{em:lineto} }
\special{em:linewidth 1.0pt}
\put(0.995,0.154){\special{em:moveto} }
\put(0.170,0.351){\special{em:lineto} }
\put(0.995,0.154){\special{em:moveto} }
\put(1.560,0.442){\special{em:lineto} }
\put(0.995,0.154){\special{em:moveto} }
\put(0.995,1.091){\special{em:lineto} }
\put(1.560,1.379){\special{em:moveto} }
\put(0.735,1.576){\special{em:lineto} }
\put(1.560,1.379){\special{em:moveto} }
\put(0.995,1.091){\special{em:lineto} }
\put(1.560,1.379){\special{em:moveto} }
\put(1.560,0.442){\special{em:lineto} }
\put(0.170,1.288){\special{em:moveto} }
\put(0.995,1.091){\special{em:lineto} }
\put(0.170,1.288){\special{em:moveto} }
\put(0.735,1.576){\special{em:lineto} }
\put(0.170,1.288){\special{em:moveto} }
\put(0.170,0.351){\special{em:lineto} }
\put(0.613,0.282){$x$}
\put(0.482,0.525){$y$}
\put(0.200,0.849){$z$}
\special{em:linewidth 0.4pt}

\put(0.895,0.895){$P$}

\put(1.143,0.784){$Q$}

\put(1.217,1.111){$R$}

\put(1.308,0.796){$P1$}

\put(1.364,0.731){$Q1$}

\put(1.449,1.056){$R1$}

\put(1.158,0.976){$P2$}

\put(1.290,0.830){$Q2$}

\put(1.257,1.124){$R2$}

\put(0.895,1.363){$n$}

\put(1.447,1.087){\special{em:moveto} }
\put(1.447,0.712){\special{em:lineto} }

\put(1.447,0.712){\special{em:moveto} }
\put(1.249,0.611){\special{em:lineto} }

\put(1.249,0.611){\special{em:moveto} }
\put(1.249,0.986){\special{em:lineto} }

\put(1.249,0.986){\special{em:moveto} }
\put(1.447,1.087){\special{em:lineto} }

\put(1.293,1.114){\special{em:moveto} }
\put(1.293,0.740){\special{em:lineto} }

\put(1.293,0.740){\special{em:moveto} }
\put(1.095,0.818){\special{em:lineto} }

\put(1.095,0.818){\special{em:moveto} }
\put(1.095,1.193){\special{em:lineto} }

\put(1.095,1.193){\special{em:moveto} }
\put(1.293,1.114){\special{em:lineto} }

\put(0.865,1.333){\special{em:moveto} }
\put(0.865,0.397){\special{em:lineto} }

\put(1.278,0.766){\special{em:moveto} }
\put(0.452,0.964){\special{em:lineto} }

\put(1.334,0.701){\special{em:moveto} }
\put(0.509,0.899){\special{em:lineto} }

\put(1.419,1.026){\special{em:moveto} }
\put(0.593,1.223){\special{em:lineto} }

\put(1.128,0.946){\special{em:moveto} }
\put(0.477,0.746){\special{em:lineto} }

\put(1.260,0.800){\special{em:moveto} }
\put(0.648,0.612){\special{em:lineto} }

\put(1.227,1.094){\special{em:moveto} }
\put(0.605,0.903){\special{em:lineto} }
\end{picture}
\caption{Second match of two views.} \label{zweiframeszwei}
\end{figure}

Let us consider the following scene in 3D, consisting of 
three parallel lines p1, q1, r1 
and three parallel lines p2, q2, r2 (see \rFig{zweiframeseins}). 
Let p1, p2 meet at P, let q1 meet q2 and r1 meet r2.
Then 
planes q1/q2 and r1/r2 are parallel. Let n be a line crossing P and
orthogonal to plane q1/q2 and hence to r1/r2. Let us rotate the body p2, q2, r2
around the n-axis. Let the rotated images be p2', q2', r2' 
(see \rFig{zweiframeszwei}). 
Then p2' still
crosses p1 at P, q2' lies in 
the plane q1/q2 - hence unless parallel to q1 it
meets q1 (say at Q') and r2' lies in r1/r2 and hence meets somewhere r1 (say
at R'). Let  a plane $\pi$  be orthogonal to p2  q2 r2.
Let P2, Q2, R2 be points of intersection of $\pi$ and p2 q2 r2 respectively.
But let us consider 
a plane $\pi'$ orthogonal to p2'q2'r2'. n is then parallel
to this plane.
Let P2', Q2', R2' be 
points of intersection of $\pi'$ and p2' q2' r2'
respectively. 
 Distances of n to p2'q2'r2' are the same as to p2q2r2. So are the
angles between image of n2 and P2'Q2', Q2'R2', R2'P2'  and the 
angles between image of n2 and P2Q2, Q2R2, R2P2. 
But r1 is orthogonal to n and so to its image in the plane P2'Q2'R2'.
Hence  the
angles between image of r1 and P2'Q2', Q2'R2', R2'P2'  and the 
angles between image of r1 and P2Q2, Q2R2, R2P2
are pairwise identical.  
 This implies that the lines on which points lie in the second image are
determined from the three points. So the forth point T2 does carry only one
piece of information in the second image instead of two. Hence there exists no
possibility of recovery of 3-D structure from two images.

To convince the reader that the difference 
between structure of the two recovered triangles PQR and PQ'R'
is meaningful, we give 
spacial coordinates of points P, Q, R, Q', R': 
P(0.0, 0.0, 0.0), 
Q(3.46537, 2.0000, -2.0000), 
Q'(4.63902, 2.0000, -2.0000), 
R(.68697, 5.0000, 4.0000), 
R'(4.37296, 5.0000, 4.0000).

\subsection{Point identity problem with two frames}

The question seems at this point to be justified what happens with the one
degree of freedom left unused. Let us consider what this freedom means
geometrically. Three traced points ensure that for every other point of the
first image we can identify the line which it lies on. This means a point T
with its image T1 in the first frame must have its image lying on a concrete
line t2 in the second frame. But if T2 does not lie on the pre-specified
line? Than two things may have happened. Either T is not a part of a rigid
body containing P, Q and R, or .... the identities of P2, Q2 and R2 have been
assigned incorrectly. 

But the latter means that if we have a set of projected points S1 and a set of
projected points S2 of which we know that they are projections of a set of
points belonging to a rigid body, but the identities are not ascribed, then
we may be capable of assigning identity relations among points of the set S1
and the set S2. For this purpose we may select four points from the set S1
and try allocating to them points of the set S2. In all, if n is the
cardinality of the set S2 (equal to the cardinality of the set S1) we may have
to try
$ n \cdot (n-1) \cdot (n-2) \cdot (n-3) $ 
combinations of points. (In case of n=4 we have 4*3*2*1=24 combinations).
First three points are then used to identify the line on which the forth point
should lie in the second frame, and the distance between the line and the real
position of projected point will be used to evaluate the goodness (or in fact
the badness) of fit. The identity assignment minimizing the distance may be
considered as the best. 

The detailed procedure has been run in an implementation as follows:
Let us denote the four traced points with P,Q,R,T and their distances as:
\\
\begin{center}
\begin{tabular}{|r|r|r|r|r|r|}
\hline 
           PQ & QR & RP & TR & TP & TQ \\
\hline 
           a  & b  & c  &  d &  f &  g \\
\hline 
\end{tabular}
\end{center}
\ \\
\newcommand{\deltab}{\Delta_b} 
 
\newcommand{\AFcbcb}{f_{cb}}
\newcommand{\AFcbc}{f_{c}}
\newcommand{\AFcbb}{f_{b}}
\newcommand{\AFcbcon}{f_{Cst}}
\newcommand{\AFcbbdwa}{f_{b ^2}}
\newcommand{\AFcbcdwa}{f_{c ^2}}

Assuming the length of c to be some number, we calculate b as a function of c
(from the triangle with edges a,b,c).
Details are given in Appendix A.
Substituting constant expressions -as in Appendix A - with constant symbols 
 $ \AFcbcb  \AFcbb \deltab,\AFcbbdwa$
 we get the simple expression:

\begin{eqnarray}  
\la 
{b}^4*\AFcbbdwa+{b}^2*({c}^2*\AFcbcb+\AFcbb)+({c}^2 * \AFcbc + {c}^4*
\AFcbcdwa + \AFcbcon)=0.\nn
\end{eqnarray}

Obviously, it is quadratic in $b{^2}$ and assuming knowledge of$c$
we introduce the notation 
$\deltab = ({c}^2*\AFcbcb+\AFcbb)^2 -4*\AFcbbdwa*({c}^2 * \AFcbc + {c}^4*
\AFcbcdwa + \AFcbcon), $
yielding:
$b  = \sqrt{
 ( -({c}^2*\AFcbcb + \AFcbb)
  \pm
  \sqrt{\deltab} ) / (2*\AFcbbdwa ).
 }
$

The a priori selected value for c can on the one hand grow towards infinity
(which may result
in the lost of precision), however it must fulfill several requirements for
its minimal value.  First of all it must be at least as long as its longest
projection. Second, it shall not lead to negative $\deltab$ and also the
resulting values of b and a as function of c shall be positive (and larger
than their projections).  \\
Then one proceeds as follows: 
Let us establish the coordinate system with axes $RP$, $RQ$, $RP\times RQ$,
assuming that point R lies in the first and in the second frame plane. Let T1
denote the projection
of the point T into the first frame, T2 - into the second.  Let us denote 
with $T_a$ the point of intersection of the plane spanned by points RPQ, and
with $T_b$ the point of intersection of the plane spanned by points RPQ and
shifted by one unit in the direction  $RP\times RQ$. It is easy to determine
then coordinates of $T_a$ and $T_b$ in he coordinate system  $RP$, $RQ$,
$RP\times RQ$. Notice that projections  $T1_a$ and $T1_b$  of points  $T_a$
and $T_b$ onto the first frame coincide with $T1$. We can now simply identify
projections  $T2_a$ and $T2_b$ of points $T_a$ and $T_b$ onto the second
frame.  Points  $T2$, $T2_a$ and $T2_b$ should be collinear in the second
frame. If it is not the case, then something is wrong about assigning
correspondences between points $P1,Q1,R1,T1$ and $P2,Q2,R2,T2$. Another
combination of assignment of point-to-point-correspondences should be tried.
(In practice, as we always run at risk of numeric rounding errors, we try to
minimize the distance between the point $T2$ and  the line  $T2_aT2_b$). 

As we have seen, b can in fact take one of two distinct values, hence the
whole procedure is to be repeated twice and the lower value of the distance of
the point $T2$ to the line $T2_aT2_b$ shall be taken.

\newcommand{\DreieckUP}[4]{}
\newcommand{\DreieckUndPunkt}[4]
{%
 \begin{picture}(10,5)    
 \put(#4){\special{em:moveto} }  
 \put(#2){\special{em:lineto} }  
 \put(#3){\special{em:lineto} }  
 \put(#4){\special{em:lineto} }  

 \put(#1){\special{em:moveto} }  
 \put(#4){\special{em:moveto} }  
 \put(#1){\special{em:lineto} }  

 \put(#1){T}
 \put(#2){Q}
 \put(#3){R}
 \put(#4){P}

 \end{picture}
%
%
}%

\newcommand{\DreieckMitGerade}[6]
{%
\begin{picture}(10,10)(-5,-5)
 \put(#4){\special{em:moveto} }  
 \put(#2){\special{em:lineto} }  
 \put(#3){\special{em:lineto} }  
 \put(#4){\special{em:lineto} }  
 \put(#1){\special{em:moveto} }  
 \put(#4){\special{em:moveto} }  
 \put(#1){\special{em:lineto} }  

 \put(#5){\special{em:moveto} }  
 \put(#6){\special{em:lineto} }  

 \put(#1){T}
 \put(#2){Q}
 \put(#3){R}
 \put(#4){P}
 \put(#5){$T_a$}
 \put(#6){$T_b$}

 \end{picture}
%
%
}%

\begin{Bsp}
Let us take the following two frames with four projected points on each.\\
Frame 1:\\
(5.301154,2.639265), 
(4.713916,1.319633),
(0.000000,0.000000),
(4.952014,0.000000).\\
Frame 2:\\
(4.509076,1.042773),
(4.642879,0.521386),
(0.000000,0.000000),
(5.732019,0.000000)
(the third point in each frame has been shifted to the origin of the
coordinate system), see the  \rFig{ZuordnungEins}.
\begin{figure}[H]
\setlength\unitlength{0.7cm} 
\noindent
\begin{center}
\vspace*{-10mm}
\begin{minipage}[t]{60mm}
\DreieckUndPunkt%
{5.301154,2.639265}%
{4.713916,1.319633}%
{0.000000,0.000000}%
{4.952014,0.000000}
\end{minipage}
\hspace*{5mm}\begin{minipage}[t]{60mm}
\DreieckUndPunkt%
{4.509076,1.042773}%
{4.642879,0.521386}%
{0.000000,0.000000}%
{5.732019,0.000000}
\end{minipage}
\end{center}
\caption{Two frames with four points each.}
\label{ZuordnungEins}
\end{figure}

Let us demonstrate several attempts to match the points in these frames 
The first two attempts are the correct ones, differing only by the sign in
the formula for calculating b,  see the \rFig{ZuordnungZwei}.\\

\begin{figure}[H]
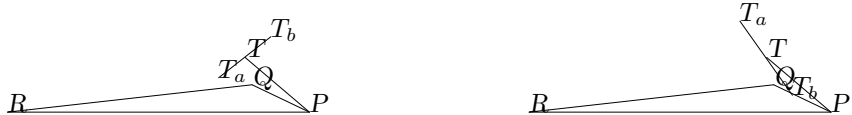

\setlength\unitlength{0.7cm} 
\noindent
\begin{center}
\vspace*{-23mm}
\Bem{
distances:
                         PQ        QR        RS          TR        TP
TQ 
            real          3   4.89898   6.40312          6   6.40312    3.4641
           real2          3   4.89898   6.40312          6   6.40312    3.4641
         squared          9        24        41         36        41        12
     projected 1     1.4444   4.89514   5.92182    4.95201   2.66226   1.34094
     projected 2   0.538281   4.67206   4.62808    5.73202   1.60716   1.20751
(gdb as db)b=   23.9807/b0=4.89701 (d=39.542,d0=6.28824) 
T1(d)(4.95201,0)1, Q1(d)(4.71392,1.31963)-1,
T2(d)(5.73202,0)1, Q2(d)(4.64288,0.521386)-1,
PaK: (-0.833333,2,0)
PbK: (13.8606,-12.3512,1)
Pa : (4.50908,1.04277,0.779541)
Pb : (20.756,13.9752,20.7072)
P2 : (4.50908,1.04277)
P1 : (5.30115,2.63927)
 (Linie) -0 = 2.29726e-14 (diff -2.29726e-14)
(version -1)
 }
\hspace*{-60mm}\begin{minipage}[t]{60mm}
\DreieckUP%
{5.301154,2.639265}%
{4.713916,1.319633}%
{0.000000,0.000000}%
{4.952014,0.000000}
\DreieckMitGerade%
{4.509076,1.042773}%
{4.642879,0.521386}%
{0.000000,0.000000}%
{5.732019,0.000000}%
{4.009076,0.652773}%
{5.000000,1.433545}
\end{minipage}
\Bem{
(gdb as db)b=   34.0514/b0=5.83536 (d=39.542,d0=6.28824) 
T1(d)(4.95201,0)1, Q1(d)(4.71392,1.31963)1,
T2(d)(5.73202,0)1, Q2(d)(4.64288,0.521386)1,
PaK: (-0.833333,2,0)
PbK: (2.03142,0.075473,1)
Pa : (4.50908,1.04277,4.83758)
Pb : (10.6464,-7.99559,8.50515)
P2 : (4.50908,1.04277)
P1 : (5.30115,2.63927)
 (Linie) -0 = -1.60554e-14 (diff 1.60554e-14)
(version 1)
 }
\hspace*{7mm}\begin{minipage}[t]{60mm}
\DreieckUP%
{5.301154,2.639265}%
{4.713916,1.319633}%
{0.000000,0.000000}%
{4.952014,0.000000}
\DreieckMitGerade%
{4.509076,1.042773}%
{4.642879,0.521386}%
{0.000000,0.000000}%
{5.732019,0.000000}%
{4.000076,1.732773}%
{5.000000,0.319794}
\end{minipage}
\end{center}
\vspace*{-35mm}
\caption{Two frames with four points each - when assignment of identities is
correcty.} \label{ZuordnungZwei}
\end{figure}

\noindent
And now some  of other conceivable assignments of correspondences. We see that
the assumed point T2 is distant from the line $T2_aT2_b$, see the 
\rFig{ZuordnungDrei}.
\end{Bsp}

%
\subsection{Linearization with 5 points}

Five points allow for a linearization as we can consider the position of
intersection of the line SS1 connecting S with its projection S1 in the first
frame with the planes PQT and QRP. Let call them $S_a$ and $S_b$ resp. The
projections of S, $S_a$, $S_b$ coincide in the first frame. Coordinates of 
$S_a$ in terms of vectors TP,~TQ and $S_b$ in terms of vectors RP,~RQ are
easily obtained from the first frame. 
Easily we can
find projections of these points onto  the second frame
(using vectors  T2P2,~T2Q2,  R2P2,~R2Q2). Let us call these projections
$S2_a$, $S2_b$.  
$S2_a$ and  $S2_b$.  
form the line which should contain the projection S2 of point S in the second
frame.  In  a  straightforward  manner  we  can   calculate   the 
discrepancy from this
assumption.

The gain form linearization in this way is that we are free from
considerations of solution requirements for quadratic equations. We have
always to solve linear equation systems only. The loss is the increased 
combinatorial 
complexity as we have to match five points instead of four.

\noindent
\begin{figure}[H]
\setlength\unitlength{0.20cm} 
\noindent
\Bem{
(gdb as db)b=   42.2572/b0=6.50055 (d=37.4966,d0=6.12345) 
T1(d)(4.95201,0)1, Q1(d)(4.71392,1.31963)1,
T2(d)(4.64288,0.521386)1, Q2(d)(5.73202,0)1,
PaK: (-0.833333,2,0)
PbK: (-2.90427,5.18389,1)
Pa : (7.59497,-0.434489,2.83364)
Pb : (17.8286,6.93936,1.40978)
P2 : (4.50908,1.04277)
P1 : (5.30115,2.63927)
 (Linie) -19.6764 = 18.1963 (diff -37.8727)
(version -1)
 (gdb as db)b=   193.476/b0=13.9096 (d=37.4966,d0=6.12345) 
T1(d)(4.95201,0)1, Q1(d)(4.71392,1.31963)-1,
T2(d)(4.64288,0.521386)1, Q2(d)(5.73202,0)-1,
PaK: (-0.833333,2,0)
PbK: (58.883,-59.7242,1)
Pa : (7.59497,-0.434489,22.0486)
Pb : (-75.5616,112.232,-526.83)
P2 : (4.50908,1.04277)
P1 : (5.30115,2.63927)
 (Linie) 118.285 = 343.118 (diff -224.833)
(version 1)
 }
\noindent
\hspace*{-10mm}
\DreieckUP%
{5.301154,2.639265}%
{4.713916,1.319633}%
{0.000000,0.000000}%
{4.952014,0.000000}
\DreieckMitGerade%
{4.509076,1.042773}%
{5.732019,0.000000}%
{0.000000,0.000000}%
{4.642879,0.521386}%
{5.000000,3.081367}%
{3.583901,5.000000}
\Bem{

(gdb as db)b=   24.1364/b0=4.91288 (d=39.0138,d0=6.2461) 
T1(d)(4.95201,0)1, Q1(d)(4.71392,1.31963)-1,
T2(d)(5.73202,0)1, Q2(d)(4.50908,1.04277)-1,
PaK: (-0.833333,2,0)
PbK: (14.6155,-13.1634,1)
Pa : (4.24147,2.08555,1.22893)
Pb : (21.8336,6.91145,20.5464)
P2 : (4.64288,0.521386)
P1 : (5.30115,2.63927)
 (Linie) 26.889 = -2.56503 (diff 29.454)
(version -1)
 }
\DreieckUP%
{5.301154,2.639265}%
{4.713916,1.319633}%
{0.000000,0.000000}%
{4.952014,0.000000}
\DreieckMitGerade%
{4.642879,0.521386}%
{4.509076,1.042773}%
{0.000000,0.000000}%
{5.732019,0.000000}%
{4.241469,2.085545}%
{5.000000,2.293627}
\Bem{
(gdb as db)b=   31.2717/b0=5.59211 (d=39.0138,d0=6.2461) 
T1(d)(4.95201,0)1, Q1(d)(4.71392,1.31963)1,
T2(d)(5.73202,0)1, Q2(d)(4.50908,1.04277)1,
PaK: (-0.833333,2,0)
PbK: (3.46802,-1.45294,1)
Pa : (4.24147,2.08555,4.20986)
Pb : (10.7397,-8.31802,10.0224)
P2 : (4.64288,0.521386)
P1 : (5.30115,2.63927)
 (Linie) 9.53646 = 3.54822 (diff 5.98824)
(version 1)
 }
\DreieckUP%
{5.301154,2.639265}%
{4.713916,1.319633}%
{0.000000,0.000000}%
{4.952014,0.000000}
\DreieckMitGerade%
{4.642879,0.521386}%
{4.509076,1.042773}%
{0.000000,0.000000}%
{5.732019,0.000000}%
{4.241469,2.085545}%
{5.000000,0.871158}
\Bem{

(gdb as db)b=   48.7377/b0=6.98124 (d=40.5885,d0=6.37091) 
T1(d)(4.95201,0)1, Q1(d)(4.71392,1.31963)1,
T2(d)(4.50908,1.04277)1, Q2(d)(5.73202,0)1,
PaK: (-0.833333,2,0)
PbK: (-3.91581,6.36025,1)
Pa : (7.70647,-0.868977,4.32178)
Pb : (22.9561,3.04368,2.225)
P2 : (4.64288,0.521386)
P1 : (5.30115,2.63927)
 (Linie) -25.462 = 7.72728 (diff -33.1893)
(version -1)
 (gdb as db)b=   154.895/b0=12.4457 (d=40.5885,d0=6.37091) 
T1(d)(4.95201,0)1, Q1(d)(4.71392,1.31963)-1,
T2(d)(4.50908,1.04277)1, Q2(d)(5.73202,0)-1,
PaK: (-0.833333,2,0)
PbK: (54.739,-55.2571,1)
Pa : (7.70647,-0.868977,18.4457)
Pb : (-81.4325,131.989,-376.747)
P2 : (4.64288,0.521386)
P1 : (5.30115,2.63927)
 (Linie) 119.676 = 402.764 (diff -283.088)
(version 1)
 }
\DreieckUP%
{5.301154,2.639265}%
{4.713916,1.319633}%
{0.000000,0.000000}%
{4.952014,0.000000}
\DreieckMitGerade%
{4.642879,0.521386}%
{5.732019,0.000000}%
{0.000000,0.000000}%
{4.509076,1.042773}%
{5.000000,3.164915}%
{3.768779,5.000000}
\Bem{

ERROR: Fdg_dg  / Fdg_g2 = -0.0278573
ERROR: Fdg_dg  / Fdg_g2 = -0.0278573

ERROR: Fdg_dg  / Fdg_g2 = -0.0685674
ERROR: Fdg_dg  / Fdg_g2 = -0.0685674

ERROR: Fdg_dg  / Fdg_g2 = -2.10003
ERROR: Fdg_dg  / Fdg_g2 = -2.10003

ERROR: Fdg_dg  / Fdg_g2 = -1.72575
ERROR: Fdg_dg  / Fdg_g2 = -1.72575

(gdb as db)b=   44.8768/b0=6.69902 (d=25.5224,d0=5.05197) 
T1(d)(4.95201,0)1, Q1(d)(4.71392,1.31963)-1,
T2(d)(1.22294,-1.04277)1, Q2(d)(-4.50908,-1.04277)1,
PaK: (-0.833333,2,0)
PbK: (19.1697,-18.7335,1)
Pa : (-10.0373,-1.21657,5.69536)
Pb : (107.858,-27.9743,-4.89544)
P2 : (0.133803,-0.521386)
P1 : (5.30115,2.63927)
 (Linie) -74.888 = 279.226 (diff -354.114)
(version -1)
 }
\DreieckUP%
{5.301154,2.639265}%
{4.713916,1.319633}%
{0.000000,0.000000}%
{4.952014,0.000000}
\DreieckMitGerade%
{0.133803,-0.521386}%
{-4.509076,-1.042773}%
{0.000000,0.000000}%
{1.222943,-1.042773}%
{-5.000000,-2.359836}%
{5.000000,-4.629454}
\Bem{
(gdb as db)b=   74.4401/b0=8.62787 (d=25.5224,d0=5.05197) 
T1(d)(4.95201,0)1, Q1(d)(4.71392,1.31963)1,
T2(d)(1.22294,-1.04277)1, Q2(d)(-4.50908,-1.04277)1,
PaK: (-0.833333,2,0)
PbK: (-22.5457,25.089,1)
Pa : (-10.0373,-1.21657,10.5718)
Pb : (-143.299,-33.1533,68.7265)
P2 : (0.133803,-0.521386)
P1 : (5.30115,2.63927)
 (Linie) 99.7118 = 331.902 (diff -232.19)
(version 1)
 
(gdb as db)b=   65.8989/b0=8.11781 (d=26.6819,d0=5.16545) 
T1(d)(4.95201,0)1, Q1(d)(4.71392,1.31963)-1,
T2(d)(-4.50908,-1.04277)1, Q2(d)(1.22294,-1.04277)1,
PaK: (-0.833333,2,0)
PbK: (27.6877,-27.5503,1)
Pa : (6.20345,-1.21657,14.0025)
Pb : (-164.444,38.5415,-149.727)
P2 : (0.133803,-0.521386)
P1 : (5.30115,2.63927)
 (Linie) 114.411 = 237.098 (diff -122.686)
(version -1)
 }
\DreieckUP%
{5.301154,2.639265}%
{4.713916,1.319633}%
{0.000000,0.000000}%
{4.952014,0.000000}
\DreieckMitGerade%
{0.133803,-0.521386}%
{1.222943,-1.042773}%
{0.000000,0.000000}%
{-4.509076,-1.042773}%
{5.000000,-0.936184}%
{-5.000000,1.393654}
\Bem{
(gdb as db)b=   1098.22/b0=33.1394 (d=26.6819,d0=5.16545) 
T1(d)(4.95201,0)1, Q1(d)(4.71392,1.31963)1,
T2(d)(-4.50908,-1.04277)1, Q2(d)(1.22294,-1.04277)1,
PaK: (-0.833333,2,0)
PbK: (-112.525,119.744,1)
Pa : (6.20345,-1.21657,64.289)
Pb : (621.699,144.529,3711.42)
P2 : (0.133803,-0.521386)
P1 : (5.30115,2.63927)
 (Linie) -432.101 = 880.406 (diff -1312.51)
(version 1)
 
ERROR: Root of deltab negative ERROR: Root of deltab negative 
(gdb as db)b=   43.4315/b0=6.59026 (d=27.3453,d0=5.22928) 
T1(d)(4.95201,0)1, Q1(d)(4.71392,1.31963)-1,
T2(d)(-4.50908,-1.04277)1, Q2(d)(0.133803,-0.521386)1,
PaK: (-0.833333,2,0)
PbK: (21.0892,-20.5595,1)
Pa : (4.02517,-0.173795,11.1078)
Pb : (-103.424,18.6706,-81.2101)
P2 : (1.22294,-1.04277)
P1 : (5.30115,2.63927)
 (Linie) -90.9355 = 55.2414 (diff -146.177)
(version -1)
 }
\DreieckUP%
{5.301154,2.639265}%
{4.713916,1.319633}%
{0.000000,0.000000}%
{4.952014,0.000000}
\DreieckMitGerade%
{1.222943,-1.042773}%
{0.133803,-0.521386}%
{0.000000,0.000000}%
{-4.509076,-1.042773}%
{4.025170,-0.173795}%
{-5.000000,1.409041}
\Bem{
(gdb as db)b=   909.231/b0=30.1535 (d=27.3453,d0=5.22928) 
T1(d)(4.95201,0)1, Q1(d)(4.71392,1.31963)1,
T2(d)(-4.50908,-1.04277)1, Q2(d)(0.133803,-0.521386)1,
PaK: (-0.833333,2,0)
PbK: (-100.956,107.65,1)
Pa : (4.02517,-0.173795,58.2687)
Pb : (439.454,185.415,3002.24)
P2 : (1.22294,-1.04277)
P1 : (5.30115,2.63927)
 (Linie) 380.812 = 522.497 (diff -141.685)
(version 1)
 }
\DreieckUP%
{5.301154,2.639265}%
{4.713916,1.319633}%
{0.000000,0.000000}%
{4.952014,0.000000}
\DreieckMitGerade%
{1.222943,-1.042773}%
{0.133803,-0.521386}%
{0.000000,0.000000}%
{-4.509076,-1.042773}%
{4.025170,-0.173795}%
{5.000000,0.241698}
\Bem{

(gdb as db)b=   35.2588/b0=5.93791 (d=44.3294,d0=6.65804) 
T1(d)(4.95201,0)1, Q1(d)(4.71392,1.31963)1,
T2(d)(1.08914,-0.521386)1, Q2(d)(-0.133803,0.521386)1,
PaK: (-0.833333,2,0)
PbK: (3.4801,-1.28541,1)
Pa : (-1.17522,1.47726,6.37058)
Pb : (-2.53473,-9.80136,15.6832)
P2 : (-4.64288,-0.521386)
P1 : (5.30115,2.63927)
 (Linie) 4.21345 = -32.1798 (diff 36.3932)
(version -1)
 }
\DreieckUP%
{5.301154,2.639265}%
{4.713916,1.319633}%
{0.000000,0.000000}%
{4.952014,0.000000}
\DreieckMitGerade%
{-4.642879,-0.521386}%
{-0.133803,0.521386}%
{0.000000,0.000000}%
{1.089140,-0.521386}%
{-1.175223,1.477261}%
{-1.955982,-5.000000}
\Bem{
(gdb as db)b=   57.7049/b0=7.59637 (d=44.3294,d0=6.65804) 
T1(d)(4.95201,0)1, Q1(d)(4.71392,1.31963)1,
T2(d)(1.08914,-0.521386)1, Q2(d)(-0.133803,0.521386)1,
PaK: (-0.833333,2,0)
PbK: (-5.26384,7.90018,1)
Pa : (-1.17522,1.47726,9.6982)
Pb : (-14.1547,-2.26527,25.8943)
P2 : (-4.64288,-0.521386)
P1 : (5.30115,2.63927)
 (Linie) -19.0107 = -6.04718 (diff -12.9635)
(version 1)
 }
\DreieckUP%
{5.301154,2.639265}%
{4.713916,1.319633}%
{0.000000,0.000000}%
{4.952014,0.000000}
\DreieckMitGerade%
{-4.642879,-0.521386}%
{-0.133803,0.521386}%
{0.000000,0.000000}%
{1.089140,-0.521386}%
{-1.175223,1.477261}%
{-5.000000,0.374413}
\Bem{

(gdb as db)b=   89.0553/b0=9.43691 (d=131.496,d0=11.4672) 
T1(d)(4.95201,0)1, Q1(d)(4.71392,1.31963)1,
T2(d)(-0.133803,0.521386)1, Q2(d)(1.08914,-0.521386)1,
PaK: (-0.833333,2,0)
PbK: (8.27237,-4.67022,1)
Pa : (2.28978,-1.47726,9.17323)
Pb : (4.65868,20.476,50.5479)
P2 : (-4.64288,-0.521386)
P1 : (5.30115,2.63927)
 (Linie) -8.89113 = 145.568 (diff -154.459)
(version -1)
 }
\DreieckUP%
{5.301154,2.639265}%
{4.713916,1.319633}%
{0.000000,0.000000}%
{4.952014,0.000000}
\DreieckMitGerade%
{-4.642879,-0.521386}%
{1.089140,-0.521386}%
{0.000000,0.000000}%
{-0.133803,0.521386}%
{2.289782,-1.477261}%
{2.988720,5.000000}
\Bem{
(gdb as db)b=   173.167/b0=13.1593 (d=131.496,d0=11.4672) 
T1(d)(4.95201,0)1, Q1(d)(4.71392,1.31963)1,
T2(d)(-0.133803,0.521386)1, Q2(d)(1.08914,-0.521386)1,
PaK: (-0.833333,2,0)
PbK: (-6.54098,10.8913,1)
Pa : (2.28978,-1.47726,16.6621)
Pb : (25.5418,5.13995,67.2957)
P2 : (-4.64288,-0.521386)
P1 : (5.30115,2.63927)
 (Linie) -28.8527 = 39.2481 (diff -68.1008)
(version 1)
 }
\DreieckUP%
{5.301154,2.639265}%
{4.713916,1.319633}%
{0.000000,0.000000}%
{4.952014,0.000000}
\DreieckMitGerade%
{-4.642879,-0.521386}%
{1.089140,-0.521386}%
{0.000000,0.000000}%
{-0.133803,0.521386}%
{2.289782,-1.477261}%
{5.000000,-0.705969}
\Bem{

(gdb as db)b=   44.8833/b0=6.6995 (d=25.5224,d0=5.05197) 
T1(d)(4.95201,0)1, Q1(d)(4.71392,1.31963)-1,
T2(d)(1.08914,-0.521386)1, Q2(d)(-4.64288,-0.521386)1,
PaK: (-0.833333,2,0)
PbK: (19.1723,-18.7361,1)
Pa : (-10.1934,-0.608284,5.5152)
Pb : (107.925,-28.2328,1.09888)
P2 : (-0.133803,0.521386)
P1 : (5.30115,2.63927)
 (Linie) -122.071 = 289.255 (diff -411.326)
(version -1)
 }
\DreieckUP%
{5.301154,2.639265}%
{4.713916,1.319633}%
{0.000000,0.000000}%
{4.952014,0.000000}
\DreieckMitGerade%
{-0.133803,0.521386}%
{-4.642879,-0.521386}%
{0.000000,0.000000}%
{1.089140,-0.521386}%
{-5.000000,-1.822866}%
{5.000000,-4.161580}
\Bem{
(gdb as db)b=   73.1452/b0=8.5525 (d=25.5224,d0=5.05197) 
T1(d)(4.95201,0)1, Q1(d)(4.71392,1.31963)1,
T2(d)(1.08914,-0.521386)1, Q2(d)(-4.64288,-0.521386)1,
PaK: (-0.833333,2,0)
PbK: (-22.2181,24.7448,1)
Pa : (-10.1934,-0.608284,10.2392)
Pb : (-140.263,-31.8954,65.2814)
P2 : (-0.133803,0.521386)
P1 : (5.30115,2.63927)
 (Linie) 158.3 = 326.099 (diff -167.799)
(version 1)
 
(gdb as db)b=   73.2025/b0=8.55585 (d=26.4894,d0=5.14678) 
T1(d)(4.95201,0)1, Q1(d)(4.71392,1.31963)-1,
T2(d)(-4.64288,-0.521386)1, Q2(d)(1.08914,-0.521386)1,
PaK: (-0.833333,2,0)
PbK: (29.3755,-29.3421,1)
Pa : (6.04735,-0.608284,15.1413)
Pb : (-171.635,41.6601,-182.124)
P2 : (-0.133803,0.521386)
P1 : (5.30115,2.63927)
 (Linie) 193.74 = 254.285 (diff -60.5448)
(version -1)
 }
\DreieckUP%
{5.301154,2.639265}%
{4.713916,1.319633}%
{0.000000,0.000000}%
{4.952014,0.000000}
\DreieckMitGerade%
{-0.133803,0.521386}%
{1.089140,-0.521386}%
{0.000000,0.000000}%
{-4.642879,-0.521386}%
{5.000000,-0.359134}%
{-5.000000,2.019740}
\Bem{
(gdb as db)b=   1341.71/b0=36.6293 (d=26.4894,d0=5.14678) 
T1(d)(4.95201,0)1, Q1(d)(4.71392,1.31963)1,
T2(d)(-4.64288,-0.521386)1, Q2(d)(1.08914,-0.521386)1,
PaK: (-0.833333,2,0)
PbK: (-125.362,133.211,1)
Pa : (6.04735,-0.608284,71.4197)
Pb : (709.165,168.232,4609.12)
P2 : (-0.133803,0.521386)
P1 : (5.30115,2.63927)
 (Linie) -801.274 = 1036.64 (diff -1837.92)
(version 1)
 
ERROR: Root of deltab negative ERROR: Root of deltab negative 
(gdb as db)b=   45.4961/b0=6.74508 (d=27.1366,d0=5.20928) 
T1(d)(4.95201,0)1, Q1(d)(4.71392,1.31963)-1,
T2(d)(-4.64288,-0.521386)1, Q2(d)(-0.133803,0.521386)1,
PaK: (-0.833333,2,0)
PbK: (21.6717,-21.1892,1)
Pa : (3.60146,1.47726,11.5271)
Pb : (-102.491,8.56135,-95.0255)
P2 : (1.08914,-0.521386)
P1 : (5.30115,2.63927)
 (Linie) -207.02 = 22.8187 (diff -229.839)
(version -1)
 }
\DreieckUP%
{5.301154,2.639265}%
{4.713916,1.319633}%
{0.000000,0.000000}%
{4.952014,0.000000}
\DreieckMitGerade%
{1.089140,-0.521386}%
{-0.133803,0.521386}%
{0.000000,0.000000}%
{-4.642879,-0.521386}%
{3.601459,1.477261}%
{-5.000000,2.051605}
\Bem{
(gdb as db)b=   1063.25/b0=32.6075 (d=27.1366,d0=5.20928) 
T1(d)(4.95201,0)1, Q1(d)(4.71392,1.31963)1,
T2(d)(-4.64288,-0.521386)1, Q2(d)(-0.133803,0.521386)1,
PaK: (-0.833333,2,0)
PbK: (-110.063,117.2,1)
Pa : (3.60146,1.47726,63.2861)
Pb : (477.129,269.556,3564.99)
P2 : (1.08914,-0.521386)
P1 : (5.30115,2.63927)
 (Linie) 951.435 = 678.52 (diff 272.915)
(version 1)
 }
\DreieckUP%
{5.301154,2.639265}%
{4.713916,1.319633}%
{0.000000,0.000000}%
{4.952014,0.000000}
\DreieckMitGerade%
{1.089140,-0.521386}%
{-0.133803,0.521386}%
{0.000000,0.000000}%
{-4.642879,-0.521386}%
{3.601459,1.477261}%
{5.000000,2.269018}
\Bem{

ERROR: Fdg_dg  / Fdg_g2 = -1.61147
ERROR: Fdg_dg  / Fdg_g2 = -1.61147

ERROR: Fdg_dg  / Fdg_g2 = -1.77834
ERROR: Fdg_dg  / Fdg_g2 = -1.77834

(gdb as db)b=   84.4386/b0=9.18905 (d=1111.28,d0=33.3358) 
T1(d)(4.95201,0)1, Q1(d)(4.71392,1.31963)-1,
T2(d)(-1.08914,0.521386)1, Q2(d)(-5.73202,0)-1,
PaK: (-0.833333,2,0)
PbK: (147.828,-144.942,1)
Pa : (-10.5564,-0.434489,-13.3974)
Pb : (666.058,-121.703,3886.74)
P2 : (-1.22294,1.04277)
P1 : (5.30115,2.63927)
 (Linie) -985.748 = 1145.65 (diff -2131.39)
(version -1)
 }
\DreieckUP%
{5.301154,2.639265}%
{4.713916,1.319633}%
{0.000000,0.000000}%
{4.952014,0.000000}
\DreieckMitGerade%
{-1.222943,1.042773}%
{-5.732019,0.000000}%
{0.000000,0.000000}%
{-1.089140,0.521386}%
{-5.000000,-1.430358}%
{5.000000,-3.222643}
\Bem{
(gdb as db)b=   3234.31/b0=56.871 (d=1111.28,d0=33.3358) 
T1(d)(4.95201,0)1, Q1(d)(4.71392,1.31963)1,
T2(d)(-1.08914,0.521386)1, Q2(d)(-5.73202,0)1,
PaK: (-0.833333,2,0)
PbK: (-82.3487,96.8613,1)
Pa : (-10.5564,-0.434489,85.4012)
Pb : (-436.021,-172.267,2740.18)
P2 : (-1.22294,1.04277)
P1 : (5.30115,2.63927)
 (Linie) 642.31 = 1617.58 (diff -975.269)
(version 1)
 
(gdb as db)b=   24.0834/b0=4.90749 (d=43.587,d0=6.60204) 
T1(d)(4.95201,0)1, Q1(d)(4.71392,1.31963)-1,
T2(d)(-5.73202,0)1, Q2(d)(-1.08914,0.521386)1,
PaK: (-0.833333,2,0)
PbK: (16.4198,-14.9022,1)
Pa : (2.5984,1.04277,6.78338)
Pb : (-79.5958,15.9274,-20.0847)
P2 : (-1.22294,1.04277)
P1 : (5.30115,2.63927)
 (Linie) 0 = 56.8791 (diff -56.8791)
(version -1)
 }
\DreieckUP%
{5.301154,2.639265}%
{4.713916,1.319633}%
{0.000000,0.000000}%
{4.952014,0.000000}
\DreieckMitGerade%
{-1.222943,1.042773}%
{-1.089140,0.521386}%
{0.000000,0.000000}%
{-5.732019,0.000000}%
{2.598403,1.042773}%
{-5.000000,2.418771}
\Bem{
(gdb as db)b=   332.713/b0=18.2404 (d=43.587,d0=6.60204) 
T1(d)(4.95201,0)1, Q1(d)(4.71392,1.31963)-1,
T2(d)(-5.73202,0)1, Q2(d)(-1.08914,0.521386)-1,
PaK: (-0.833333,2,0)
PbK: (77.9445,-79.5346,1)
Pa : (2.5984,1.04277,33.671)
Pb : (-361.863,-149.361,-1195.22)
P2 : (-1.22294,1.04277)
P1 : (5.30115,2.63927)
 (Linie) 0 = -574.745 (diff 574.745)
(version 1)
 }
\DreieckUP%
{5.301154,2.639265}%
{4.713916,1.319633}%
{0.000000,0.000000}%
{4.952014,0.000000}
\DreieckMitGerade%
{-1.222943,1.042773}%
{-1.089140,0.521386}%
{0.000000,0.000000}%
{-5.732019,0.000000}%
{2.598403,1.042773}%
{-5.000000,-2.092893}
\Bem{

(gdb as db)b=   38.2674/b0=6.18607 (d=182.468,d0=13.5081) 
T1(d)(4.95201,0)1, Q1(d)(4.71392,1.31963)-1,
T2(d)(-1.22294,1.04277)1, Q2(d)(-5.73202,0)-1,
PaK: (-0.833333,2,0)
PbK: (58.7612,-57.0864,1)
Pa : (-10.4449,-0.868977,-6.52429)
Pb : (252.933,-18.4489,661.293)
P2 : (-1.08914,0.521386)
P1 : (5.30115,2.63927)
 (Linie) -353.183 = 177.482 (diff -530.665)
(version -1)
 }
\DreieckUP%
{5.301154,2.639265}%
{4.713916,1.319633}%
{0.000000,0.000000}%
{4.952014,0.000000}
\DreieckMitGerade%
{-1.089140,0.521386}%
{-5.732019,0.000000}%
{0.000000,0.000000}%
{-1.222943,1.042773}%
{-5.000000,-1.232414}%
{5.000000,-1.899892}
\Bem{
(gdb as db)b=   529.981/b0=23.0213 (d=182.468,d0=13.5081) 
T1(d)(4.95201,0)1, Q1(d)(4.71392,1.31963)1,
T2(d)(-1.22294,1.04277)1, Q2(d)(-5.73202,0)1,
PaK: (-0.833333,2,0)
PbK: (-35.1042,41.5201,1)
Pa : (-10.4449,-0.868977,33.4158)
Pb : (-171.814,-86.2172,460.9)
P2 : (-1.08914,0.521386)
P1 : (5.30115,2.63927)
 (Linie) 237.369 = 811.507 (diff -574.138)
(version 1)
 
(gdb as db)b=   24.0047/b0=4.89945 (d=43.9665,d0=6.63072) 
T1(d)(4.95201,0)1, Q1(d)(4.71392,1.31963)-1,
T2(d)(-5.73202,0)1, Q2(d)(-1.22294,1.04277)1,
PaK: (-0.833333,2,0)
PbK: (16.0699,-14.5226,1)
Pa : (2.3308,2.08555,6.47902)
Pb : (-77.8284,7.30974,-19.6281)
P2 : (-1.08914,0.521386)
P1 : (5.30115,2.63927)
 (Linie) -120.032 = 23.2157 (diff -143.248)
(version -1)
 }
\DreieckUP%
{5.301154,2.639265}%
{4.713916,1.319633}%
{0.000000,0.000000}%
{4.952014,0.000000}
\DreieckMitGerade%
{-1.089140,0.521386}%
{-1.222943,1.042773}%
{0.000000,0.000000}%
{-5.732019,0.000000}%
{2.330796,2.085545}%
{-5.000000,2.563313}
\Bem{
(gdb as db)b=   315.671/b0=17.7671 (d=43.9665,d0=6.63072) 
T1(d)(4.95201,0)1, Q1(d)(4.71392,1.31963)-1,
T2(d)(-5.73202,0)1, Q2(d)(-1.22294,1.04277)-1,
PaK: (-0.833333,2,0)
PbK: (76.3463,-77.8435,1)
Pa : (2.3308,2.08555,32.6109)
Pb : (-345.896,-186.673,-1128.88)
P2 : (-1.08914,0.521386)
P1 : (5.30115,2.63927)
 (Linie) -539.333 = -640.195 (diff 100.862)
(version 1)
 }
\DreieckUP%
{5.301154,2.639265}%
{4.713916,1.319633}%
{0.000000,0.000000}%
{4.952014,0.000000}
\DreieckMitGerade%
{-1.089140,0.521386}%
{-1.222943,1.042773}%
{0.000000,0.000000}%
{-5.732019,0.000000}%
{2.330796,2.085545}%
{-5.000000,-1.888171}
\Bem{

}
\caption{Two frames with four points each - when assigment of identities is
not correct} \label{ZuordnungDrei}
\end{figure}

\subsection{Recovery of curves from two frames}

So far we have considered only finite sets of points. We have just
demonstrated that for finite sets of tracable points two frames are
insufficient for recovery of structural and motion parameters. But can a
break-through be achieved if we have to do with infinite sets of points. Here
we must carefully outline the frontier between various classes of tasks. If
we have to do with surfaces, with shadows, textures etc., there exist various
approaches handling the problem. But we want to concentrate here on objects
where surface cannot be percepted (or percepted properly) - on curves. Objects
consisting of wires can be viewed as such objects under some circumstances.
Further let us assume that we are not aware of the category the object belongs
to. If we knew it is a circle or somme opther object with weel-studied
invariant properties, we could be helped just with this external information.
But let us assume that the object we actually traced is a true any-shaped
smooth 3-D curve (or a combination of such curves). Furthermore let us assume
we were able to establish point-to-point correspomndence of all~(!!!) points
of both curves. Let us then assuma we made a guess of the intrinsic shape of
the projected object so that it fits both of them.

 Can we enjoy our guess? The
answer is: No. We can take the very same procedure as in the first subsection
of this section (with rotation around a properly selected straight line) and
obtain another 3-D curve fiiiting both of our projections , usually totally
different from the first one. In fact, we can rotate and rotate and obtain an
infinite family of such curves, each of them fitting both projections. This is
also true of any imperfect fit. Assume, due to some digitalization (much or
less random) errors we cannot fit  both images, but we run an approximating
procedure yielding a best fit. If then we cannot attribute errors of the
digital image to tthe distance between the image and the projected object,
then we have no way to achieve unique identification of the 3-D curve. 

\section{Discussion}

In this paper we have demonstrated that for orthogonal projection of rigid
point bodies at least three frames and three points are necessary to recover
structure and motion. It has been shown that the complexity of the task is
that of solving a quadratic equation in one variable with postprocessing to
meet physical constraints. The problem under consideration may be linearized,
however, if we can trace four points over three frames or three points over
four frames.

From a degrees-of-freedom argument it became visible that the amount of
information that two frames with four traced points may provide enough
information to recover structure and motion from two frames. However, it has
been demonstrated that this is impossible because the rigid body assumption
imposes internal dependence between the point projections so that 
information
provided by the forth point and any further traced point cannot be
consumed for purposes of recovery of structure and motion. This result extends
to infinite sets of points: to objects constructed of a set of smooth true 3-D
curves.
  
Instead,  four points over two frames may solve
identification problem of points between consecutive frames or alternatively
the problem of belonging to the same rigid body. That is, in the first case,
if we have two frames with four (or more) points each and we know that these
points belong to the same rigid body, but we do not know the exact point to
point correspondence, then we can exploit the unused information (not
consumable for recovery of structure and motion) for purposes of
identification of point-to-point correspondences. Alternatively, in the second
case, when we have sets of points in two frames where the
point-to-point-correspondence between frames is known, then we can exploit the
 unused information (not
consumable for recovery of structure and motion) to decide, which points
belong to the same rigid body.
We have seen also that a fifth point can linearize the otherwise quadratic
task.

It is worth mentioning at this point that several papers claimed for analogous
problem under perspective projection that structure and motion can be
recovered from two frames when 9, 7 or 5 points,
\Zitat{R:1,R:2,fivepoints}, 
  or 4 points and 1 line,  \Zitat{linehelps}, 
 are
traced. In \Zitat{Klopotek:94c} the four- point-and-a-line
claim of  \Zitat{linehelps}
has been rejected on the basis of degrees-of-freedom argument and also
by explicit construction. It seems to be worth investigating whether the 9, 7,
and 5 points claims are valid or not, that is whether the internal constraints
of a rigid body do not exhibit same properties as in case of four points under
orthogonal projection. 

\section{Conclusions}

\begin{itemize}
\item For orthogonal projection of rigid
point bodies at least three frames and three points are necessary to recover
structure and motion, when the motion is arbitrary (under no control of the
observer).
\item Solution to the structure and motion problem under these circumstances
requires solving a quadratic equation, yielding sometimes two feasible
solutions.
\item If four frames or four points are available, then the problem can be
linearized, giving unique solution (if the motion pattern is not 
degenerate).
\item It is impossible to recover structure or motion from two frames whatever
number of traced points is available. The result extends to infinite sets of
points in case the traced object consists of smooth 3D curves (but has no
surfaces).
\item If a rigid  body consists of at least four points, then 
we can solve the problem of point tracing for any two consecutive frames alone
from knowledge which points of two frames belong to the body (without explicit
knowledge of point-to-point correspondence)
\item Alternatively, if a rigid  body consists of at least four points, then
we can solve the problem of belonging to a rigid body for any
two consecutive frames alone
from  explicit
knowledge of point-to-point correspondence.
\item The solution of the above two problems involves solution of quadratic
equation, but may be linearized if five points are available instead.
\item It is worth investigating whether these results extend to perspective
projections.

\end{itemize}

%
\newpage
\section*{Appendix A}
%
%
\newcommand{\Fcbcb}{
2* 
(
(\facc{2}-\facc{1}) \nn\\
\la
*\frac{
  (\facb{2}-\facb{1})
     }{ 
   (-\faca{2}+\faca{1})^2
}
\nn\\
\la
- 
\frac{
 (\facc{2}-\facc{1})
     }{ 
   (-\faca{2}+\faca{1})
} 
\nn\\
\la
- 
\frac{
 (\facb{2}-\facb{1})
     }{ 
   (-\faca{2}+\faca{1})
}
-1)
}
\newcommand{\Fcbc}{
\frac{
   (\facc{2}-\facc{1}) * \faca{1}
     }{ 
   (-\faca{2}+\faca{1})
} \nn\\
\la
+ \facc{1}\nn\\
\la
-2* (\facCst{2} \nn\\
\la
  -\facCst{1})\nn\\
\la
* \frac{1
     }{ 
   (-\faca{2}+\faca{1})
} \nn\\
\la
+ 2*  (\facCst{2} \nn\\
\la
 -\facCst{1})\nn\\
\la
\frac{
(\facc{2}-\facc{1})     
     }{ 
   (-\faca{2}+\faca{1})^2
}
}
\newcommand{\Fcbb}{
\frac{
   (\facb{2}-\facb{1}) * \faca{1}
     }{ 
   (-\faca{2}+\faca{1})
} \nn\\
\la
+ \facb{1}\nn\\
\la
-2*   (\facCst{2} \nn\\
\la
 -\facCst{1})\nn\\
\la
* \frac{1
     }{ 
   (-\faca{2}+\faca{1})
} \nn\\
\la
+ 2*  (\facCst{2}\nn\\
\la
  -\facCst{1})\nn\\
\la
*\frac{
(\facb{2}-\facb{1}) 
     }{ 
   (-\faca{2}+\faca{1})^2
}
}
\newcommand{\Fcbcon}{
(   
(\facCst{2} \nn\\
\la
 -\facCst{1})\nn\\
\la
*\frac{  1  }
     {    (-\faca{2}+\faca{1}) }
)^2 \nn\\
\la
+   
(\facCst{2} \nn \\
\la
 -\facCst{1})\nn\\
\la
*\frac{   \faca{1}}
     {   (-\faca{2}+\faca{1}) }\nn\\
\la
+\facCst{1}
}
\newcommand{\Fcbbdwa}{
 (1-
\frac{
 (\facb{2}-\facb{1})
    }{ 
   (-\faca{2}+\faca{1})
}
)^2
}
\newcommand{\Fcbcdwa}{
 (1-
\frac{
 (\facc{2}-\facc{1})
    }{ 
   (-\faca{2}+\faca{1})
}
)^2
}
Let us denote the four traced points with P,Q,R,T and their distances as:
\\
\begin{center}
\begin{tabular}{|r|r|r|r|r|r|}
\hline 
           PQ & QR & RP & TR & TP & TQ \\
\hline 
           a  & b  & c  &  d &  f &  g \\
\hline 
\end{tabular}
\end{center}
\ \\
Assuming the length of c can be some number, we can calculate b as a
function of c (from the triangle with edges a,b,c). We take an equation system
of the form given by equation \rEq{p3f3sqr}, with i=1,2. \\
Both equations can be considered as quadratic in ${a}^2$. In the first step
we subtract both equations getting an equation linear in ${a}^2$. We calculate
 ${a}^2$ from this equation and substitute the result into one of the original
equations just eliminating the variable $a$. The result is an equation which
is quadratic both
in ${b}^2$ and in ${c}^2$. It can be expressed in a simple manner if we adopt
 the denotation:
\begin{eqnarray}  
\la 
 {\AFcbcb}=   \Fcbcb ,     \nn 
\end{eqnarray}

\smallskip
\begin{eqnarray}  
\la 
 {\AFcbc}=   \Fcbc ,       \nn
\end{eqnarray}

\smallskip
\begin{eqnarray}  
\la 
 {\AFcbb}=   \Fcbb ,      \nn
\end{eqnarray}

\smallskip
\begin{eqnarray}  
\la 
 {\AFcbcon}=   \Fcbcon ,     \nn
\end{eqnarray}

\smallskip
\begin{eqnarray}  
\la 
 {\AFcbbdwa}=   \Fcbbdwa ,    \nn
\end{eqnarray}

\smallskip
\begin{eqnarray}  
\la 
 {\AFcbcdwa}=   \Fcbcdwa .    \nn 
\end{eqnarray}

\medskip
\noindent
Substituting constant expressions with constant symbols 
 $ \AFcbcb  \AFcbb \deltab,\AFcbbdwa$, 
 we get the simple expression:
${b}^4*\AFcbbdwa+{b}^2*({c}^2*\AFcbcb+\AFcbb)+({c}^2 * \AFcbc + {c}^4*
\AFcbcdwa + \AFcbcon)=0.$
Obviously, it is quadratic in $b{^2}$ and assuming knowledge of$c$
we introduce the notation 
$\deltab = ({c}^2*\AFcbcb+\AFcbb)^2 -4*\AFcbbdwa*({c}^2 * \AFcbc + {c}^4*
\AFcbcdwa + \AFcbcon)
$
yielding: 
$b  = \sqrt{
 ( -({c}^2*\AFcbcb + \AFcbb)
  \pm
  \sqrt{\deltab} ) / (2*\AFcbbdwa )
 }.
$

\end{document}